\documentclass[a4paper]{article}

\usepackage{INTERSPEECH2021}
\usepackage{amsmath,graphicx}
\usepackage[backend=bibtex,citestyle=numeric-comp,bibstyle=ieee,sorting=nty,defernumbers=true,giveninits=true,doi=false,isbn=false,url=false,eprint=false,minbibnames=2,maxbibnames=2]{biblatex}
\def\etal{{\em et al.~}}
\def\etal{{\em et al.~}}
\AtEveryBibitem{
    \clearlist{address}
    \clearfield{date}
    \clearfield{eprint}
    \clearfield{isbn}
    \clearfield{issn}
    \clearlist{location}
    \clearfield{month}
    \clearfield{series}
    \ifentrytype{journal}{}{
      \clearlist{publisher}
      \clearname{editor}
     }
}
\usepackage[aboveskip=2pt]{subcaption}
\usepackage{amssymb}
\usepackage{tabularx}
\usepackage{booktabs}
\usepackage{multirow}
\usepackage{ragged2e}
\usepackage{bm}

\newcolumntype{C}{>{\Centering\arraybackslash}X}
\newcolumntype{e}{>{\Centering\hsize=\hsize}X}
\newcolumntype{t}{>{\Centering\hsize=.6\hsize}X}
\newcolumntype{s}{>{\Centering\hsize=.5\hsize}X}
\newcolumntype{j}{>{\Centering\hsize=.4\hsize}X}
\newcolumntype{k}{>{\Centering\hsize=.3\hsize}X}
\newcolumntype{y}{>{\Centering\hsize=.2\hsize}X}
\usepackage{hyperref}

\usepackage{xcolor}

\makeatletter
\newcommand\thefontsize{The current font size is: \f@size pt}
\makeatother
\usepackage[normalem]{ulem}
\usepackage[activate]{microtype}
\title{LiRA: Learning Visual Speech Representations from Audio through Self-supervision}

\name{Pingchuan Ma$^{1*}$,
Rodrigo Mira$^{1*}$,
Stavros Petridis$^{2}$, 
Björn W.\ Schuller$^{1,3}$ and
Maja Pantic$^{1,2}$\thanks{$^{*}$ denotes equal contribution to this work.}}
\address{$^{1}$iBUG Group, Imperial College London, UK\\
$^{2}$Facebook London, UK \\
$^{3}$Chair of Embedded Intelligence for Health Care and Wellbeing, University of Augsburg, Germany}
\author{\IEEEauthorblockN{}
\IEEEauthorblockA{\IEEEauthorrefmark{1}IBUG Group, Imperial College London,London, UK}
\IEEEauthorblockA{\IEEEauthorrefmark{2}Facebook London, London, UK}
\IEEEauthorblockA{\IEEEauthorrefmark{3}Chair of Embedded Intelligence, for Health Care and Wellbeing, University of Augsburg, Augsburg, Germany}}
\email{pm4115@ic.ac.uk, rs2517@ic.ac.uk}

\begin{document}

\maketitle

\begin{abstract}
The large amount of audiovisual content being shared online today has drawn substantial attention to the prospect of audiovisual self-supervised learning. Recent works have focused on each of these modalities separately, while others have attempted to model both simultaneously in a cross-modal fashion. However, comparatively little attention has been given to leveraging one modality as a training objective to learn from the other. In this work, we propose Learning visual speech Representations from Audio via self-supervision (LiRA). Specifically, we train a ResNet+Conformer model to predict acoustic features from unlabelled visual speech. We find that this pre-trained model can be leveraged towards word-level and sentence-level lip-reading through feature extraction and fine-tuning experiments. We show that our approach significantly outperforms other self-supervised methods on the Lip Reading in the Wild (LRW) dataset and achieves state-of-the-art performance on Lip Reading Sentences 2 (LRS2) using only a fraction of the total labelled data.
\end{abstract}

\section{Introduction}
\label{sec:intro}

Self-supervised learning aims to leverage unlabelled data by extracting the training objective directly from the input itself, in an attempt to model meaningful representations of the proposed modality which capture its content and structure. In works adopting this methodology, this task is usually known as the ``pretext task'' and this initial training procedure is known as the ``pre-training'' stage. After pre-training, the network is trained on the ``downstream task'', which generally involves a smaller set of manually labelled data. This methodology has received substantial attention in recent years within the computer vision community.
Pretext tasks for visual self-supervision include image colourisation~\cite{DBLP:conf/eccv/ZhangIE16}, jigsaw puzzle solving~\cite{DBLP:conf/eccv/NorooziF16}, as well as combinations of these and other tasks~\cite{DBLP:conf/cvpr/KolesnikovZB19}. Self-supervised learning has also been explored in the speech community through works such as Contrastive Predicting Coding (CPC)~\cite{DBLP:journals/corr/abs-1807-03748} and wav2vec~\cite{DBLP:conf/interspeech/SchneiderBCA19}, which predict/discriminate future segments of audio samples; LIM (Local Info Max)~\cite{DBLP:conf/interspeech/RavanelliB19}, which maximises mutual information for the same speaker; and, more recently, PASE (Problem Agnostic Speech Encoder)~\cite{DBLP:conf/interspeech/PascualRSBB19,DBLP:conf/icassp/RavanelliZPSMTB20}, which predicts established audio features such as STFT and MFCC.

Self-supervision has also been adopted in the audiovisual domain. Recent approaches include audiovisual fusion~\cite{DBLP:journals/taffco/PetridisP16, petridis2011prediction}, clustering~\cite{DBLP:journals/corr/abs-1911-12667}, and distillation~\cite{DBLP:conf/cvpr/PiergiovanniAR20}; cross-modal discrimination~\cite{DBLP:conf/eccv/OwensE18}; cyclic translation between modalities~\cite{DBLP:conf/aaai/PhamLMMP19}; and permutative predictive coding \cite{DBLP:conf/icpr/TellamekalaVPG20}. Shukla \etal~\cite{DBLP:journals/corr/abs-2005-01400} focus on learning audio representations by facial reconstruction from waveform speech. Conversely,~\cite{DBLP:journals/ijcv/OwensWMFT18} predict frequency-based summaries of ambient sound from video, while other recent works apply audio-visual synchronisation~\cite{DBLP:conf/iccv/ArandjelovicZ17,DBLP:conf/nips/KorbarTT18,9067055} to learn visual embeddings. A task that can benefit from self-supervised learning is lip-reading. Current state-of-the-art lip-reading models rely on annotating hundreds of hours of visual speech data~\cite{makino2019recurrent}, which is costly. To solve this issue, Afouras \etal~\cite{DBLP:conf/icassp/AfourasCZ20} propose using a pre-trained Automatic Speech Recognition (ASR) model to produce machine-generated captions for unsupervised pre-training. This provides automatically labelled data but still relies on an ASR model trained on large amounts of labelled data.

In this work, we aim to leverage the vast amount of available audiovisual speech data to learn generic visual speech features and improve state-of-the-art lip-reading models by predicting audio features from visual speech. The targeted audio features are extracted from waveform audio without the need for additional labels using an established speech encoder (PASE+~\cite{DBLP:conf/icassp/RavanelliZPSMTB20}). Using the proposed approach, the learnt visual features are explicitly guided by audio which contains rich information about speech. This in turn can lead to learning visual features which are more suitable for speech recognition. After this training procedure, we apply our model (Fig.~\ref{fig:variations}) for lip-reading on a transcribed visual speech dataset. 

Our research contributions are as follows:

\textbf{1)} We present LiRA, which learns powerful visual speech representations by predicting acoustic features from raw video taken from large audio-visual datasets.
\textbf{2)} We demonstrate that LiRA provides a good initialisation for fine-tuning lip-reading models which consistently outperforms training from scratch, and that this method is particularly beneficial for smaller labelled datasets.
\textbf{3)} We show that LiRA outperforms previous self-supervised methods for word-level lip-reading, achieving an accuracy of 88.1\% on LRW by pre-training on unlabelled data.
\textbf{4)} Finally, we leverage our self-supervised approach towards sentence-level lip-reading, and find that our fine-tuned model achieves state-of-the-art performance for LRS2.

\vspace{-1mm}\looseness - 1
\section{Methodology}
\label{sec:method}
\begin{figure}
\centering 
\includegraphics[width=.78\linewidth]{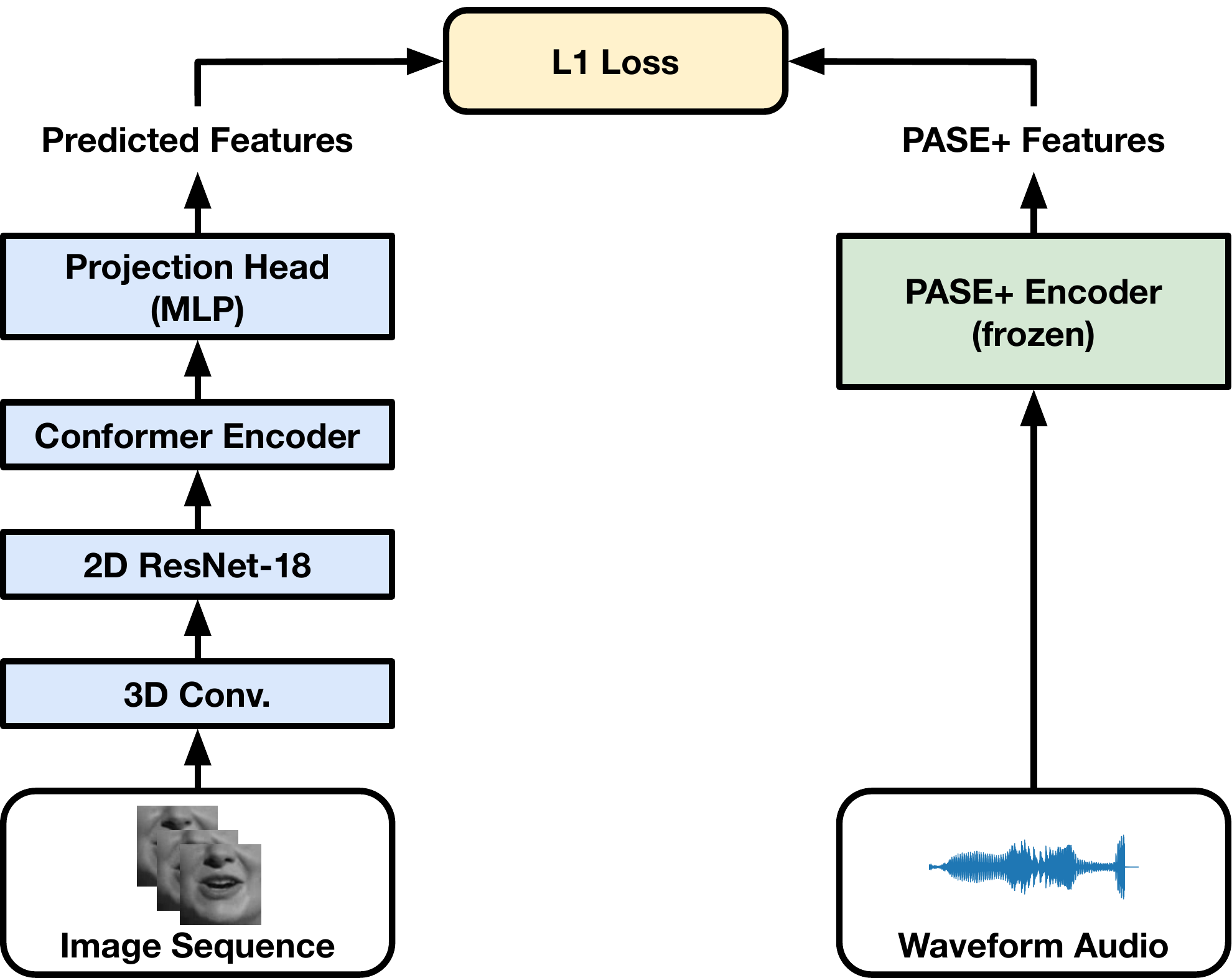}

\caption{The high-level architecture of our model and our methodology for audiovisual self-supervised training.}
\label{model}
\vspace{-7mm}
\end{figure}

\vspace{-1mm}
\subsection{Pretext task}
LiRA predicts PASE+ features from raw video and is composed of three distinct components. The first is the spatial encoder, which is a traditional 2D ResNet-18 preceded by a 3D front-end layer. The second component is the temporal encoder -- the conformer -- which receives as input the frame-wise features produced by the spatial encoder and returns a set of features of the same size. The conformer encoder combines traditional attention-based transformer blocks, which excel at capturing global temporal dependencies, with convolutional layers, which model local patterns efficiently~\cite{gulati2020conformer}. The final component is the projection head (based on the MLP -- Multi-Layer Perceptron -- workers presented in~\cite{DBLP:conf/interspeech/PascualRSBB19}), which projects these representations into the predicted PASE+ features. To train the model, we apply an L1 loss between the generated embeddings and the features extracted from the pre-trained (frozen) PASE+ model, as shown in Fig.~\ref{model}. We would also like to mention that we have also experimented with predicting MFCC features but the results were worse than predicting PASE+ features.

\vspace{-1mm}
\subsection{Downstream task}
To evaluate the visual speech representations, we run three variations of end-to-end lip-reading experiments. The training procedure is illustrated in Fig.~\ref{fig:variations}. LiRA-Supervised models are trained from scratch based on the same encoder as in the self-supervised training~\cite{ma2020large}. This serves as our baseline model since it is trained only with the labelled training data. LiRA-Frozen models are trained using LiRA features from the pre-trained encoder. This allows us to evaluate the visual representations learned during self-supervised learning. Finally, LiRA-FineTuned models use the same model as LiRA-Supervised but are initialised with the pre-trained encoder weights from the pretext task. By using this configuration, we can evaluate the model initialisation capabilities of the proposed self-supervised learning approach. For each of these methods, we adopt a separate model for each lip-reading task - six models in total. For word-level lip-reading, we use a Multi-Scale Temporal Convolutional Network (MS-TCN)~\cite{martinez2020lipreading} on top of the encoder, followed by a linear classifier for classification. For sentence-level lip-reading, we follow the state-of-the-art lip-reading model~\cite{ma2020large} on LRS2 and build a hybrid CTC/attention model. We use the same conformer encoder architecture as in the pre-training phase, followed by the transformer decoder for sequence-to-sequence training~\cite{DBLP:conf/nips/VaswaniSPUJGKP17}. We also perform fine-tuning experiments using the pre-trained model.

\vspace{-1mm}\looseness - 1
\section{Experimental Setup}

\begin{table}[!t]
\scriptsize
\centering
\begin{tabularx}{\columnwidth}{ m{0.3em} | e | e | e }
& LiRA-Supervised & LiRA-Frozen  & LiRA-FineTuned  \\
\hline
\rotatebox[origin=c]{90}{Word-level lip-reading}
&
\begin{subfigure}{.28\columnwidth}
    \includegraphics[width=\linewidth]{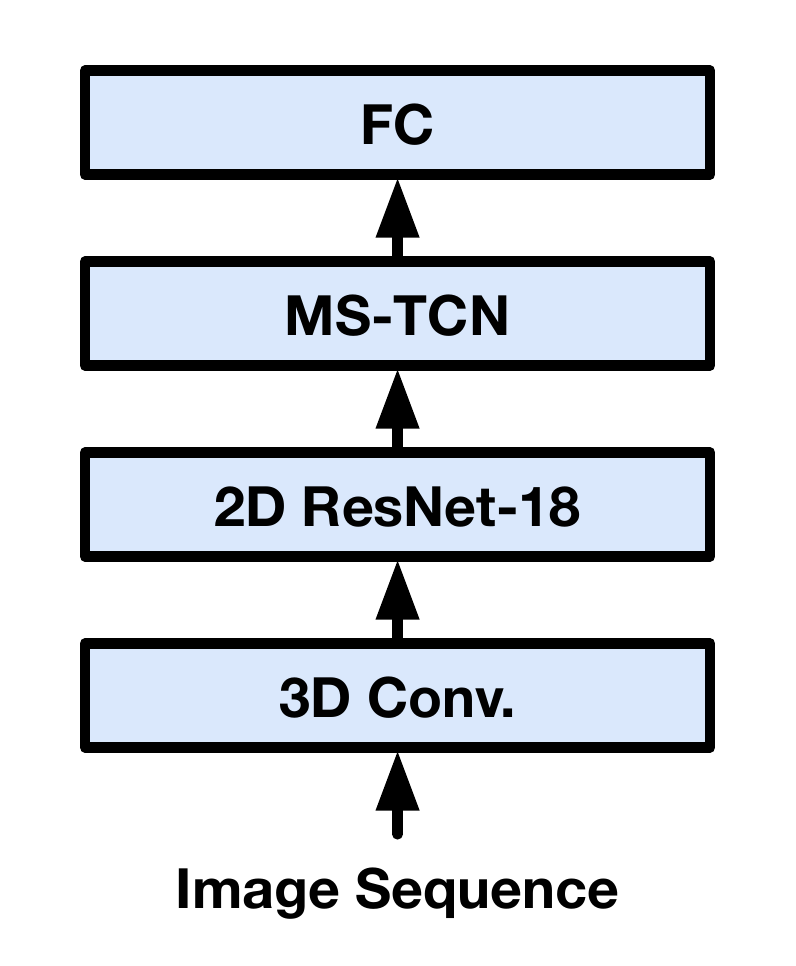}
    \caption{}
    \label{fig:variations_a}  
\end{subfigure}
&
\begin{subfigure}{.28\columnwidth}
    \includegraphics[width=\linewidth]{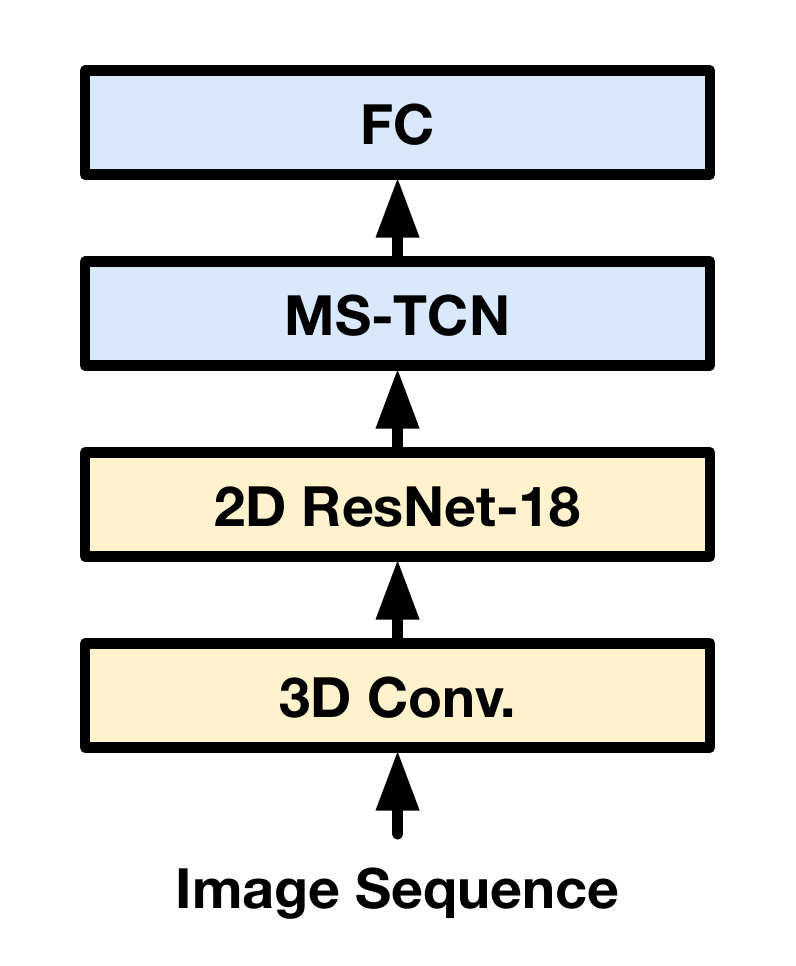}
    \caption{}
    \label{fig:variations_b}  
\end{subfigure}
&
\begin{subfigure}{.28\columnwidth}
    \includegraphics[width=\linewidth]{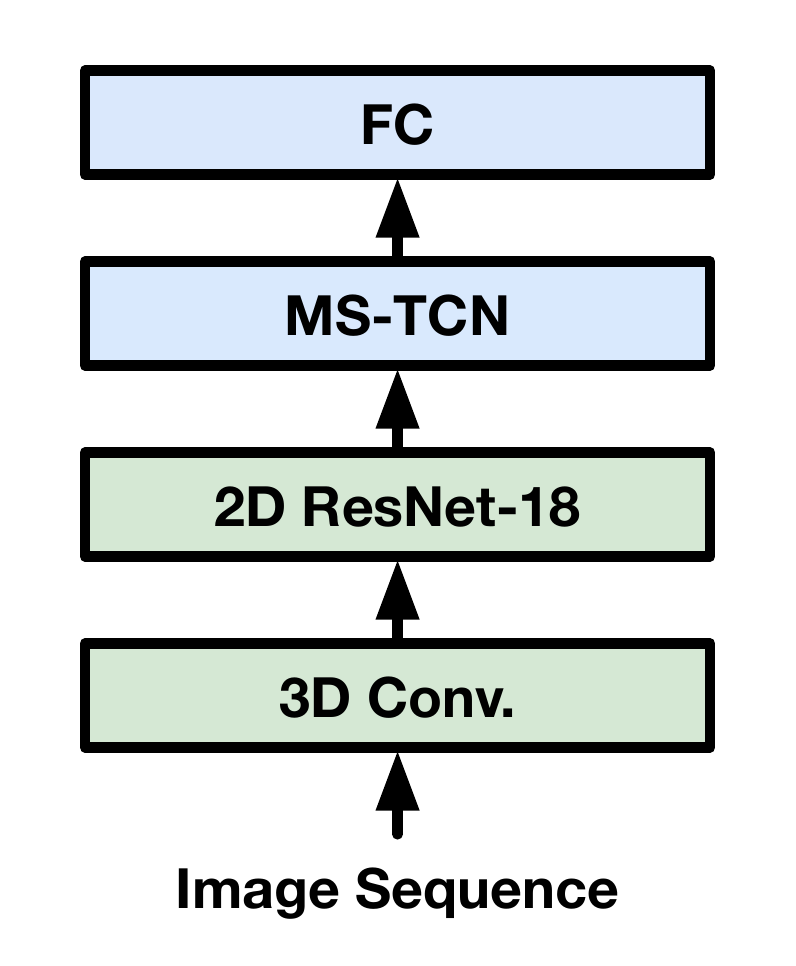}
    \caption{}
    \label{fig:variations_c}  
\end{subfigure}
\\ \hline
\rotatebox[origin=c]{90}{Sentence-level lip-reading}
&
\begin{subfigure}{.28\columnwidth}
    \includegraphics[width=\linewidth]{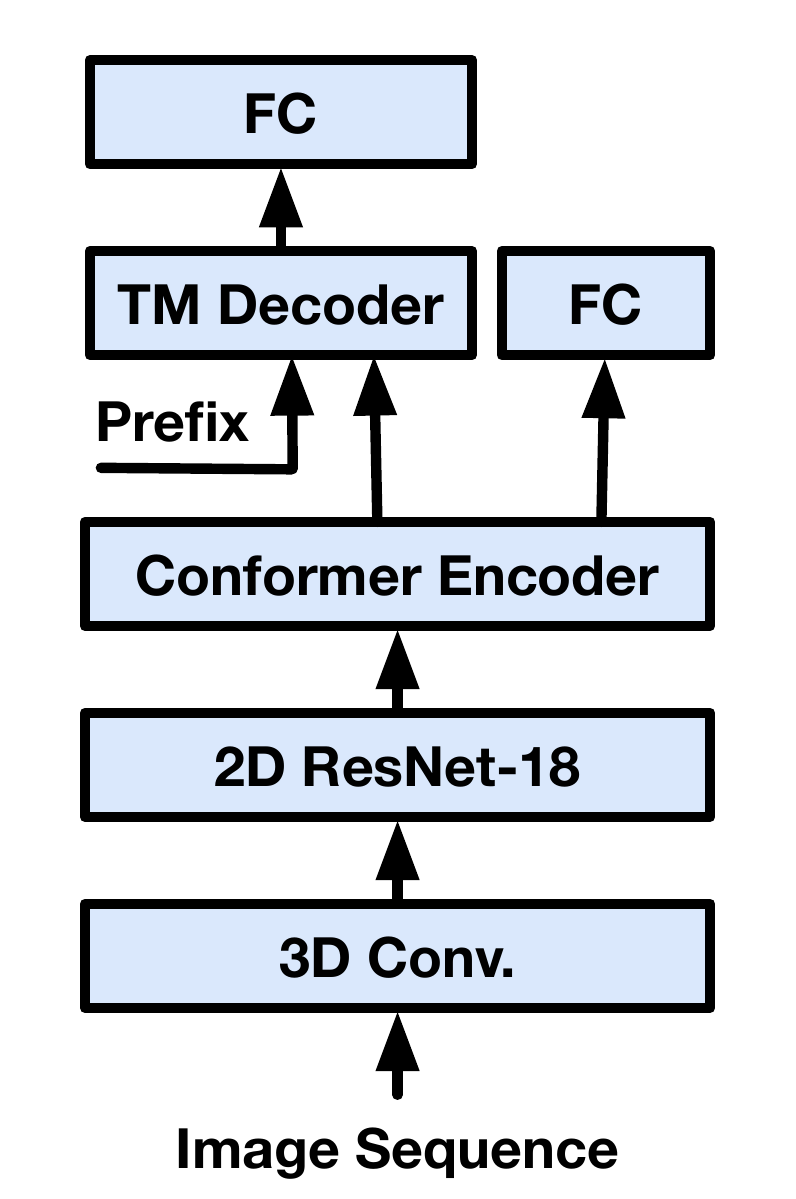}
    \caption{}
    \label{fig:variations_d}  
\end{subfigure}
&
\begin{subfigure}{.28\columnwidth}
    \includegraphics[width=\linewidth]{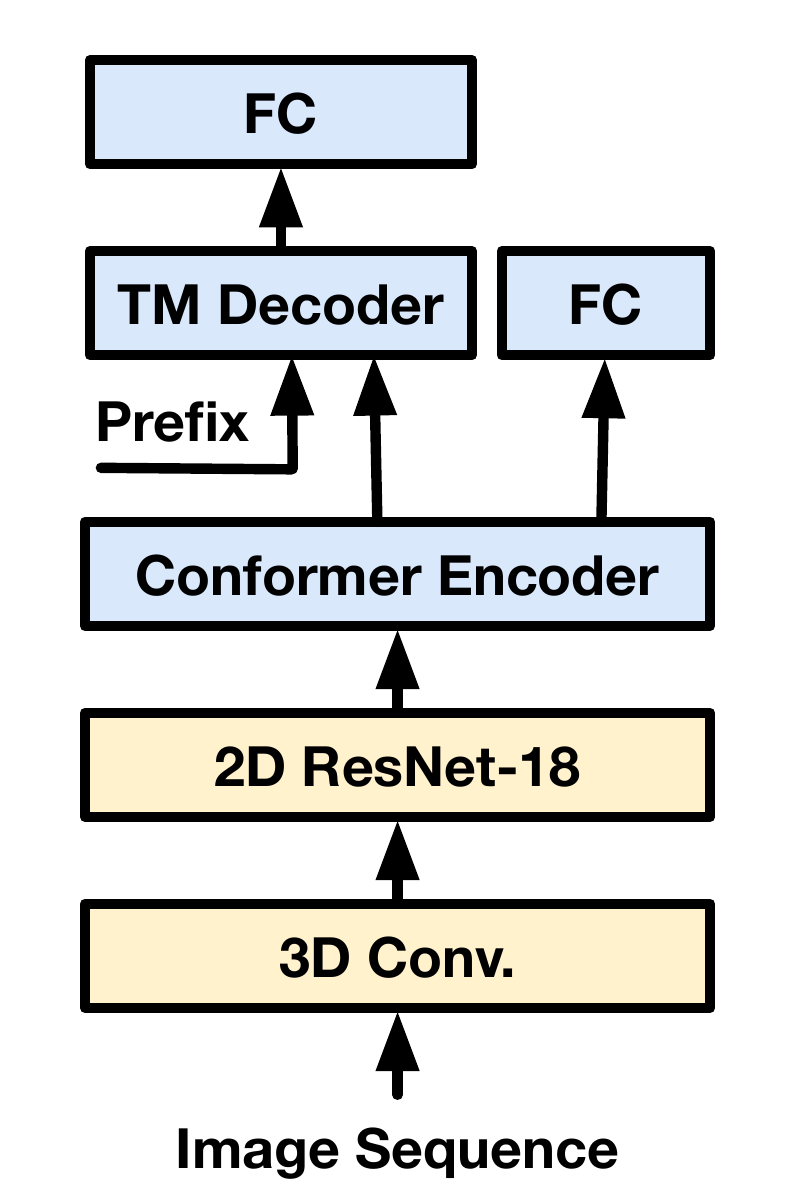}
    \caption{}
    \label{fig:variations_e}  
\end{subfigure}
&
\begin{subfigure}{.28\columnwidth}
    \includegraphics[width=\linewidth]{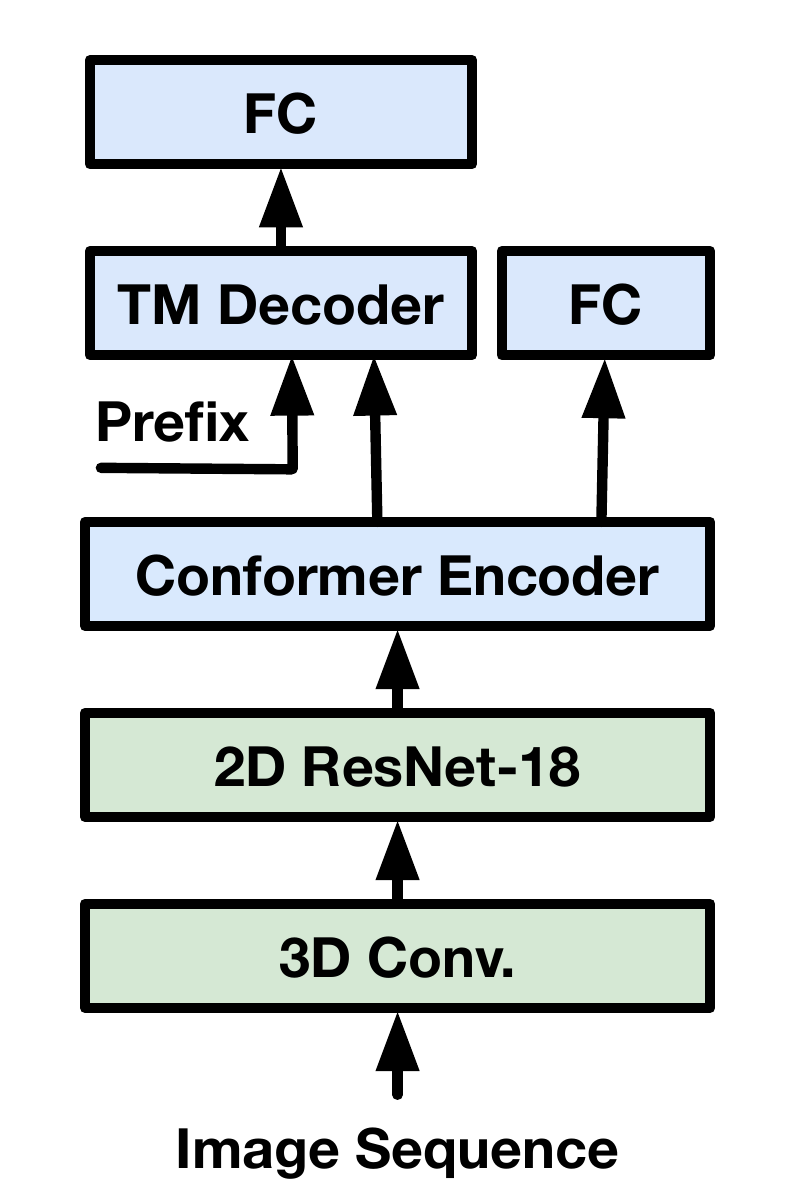}
    \caption{}
    \label{fig:variations_f}  
\end{subfigure}
\\
\end{tabularx}
\vspace{1mm}
\captionof{figure}{The variations of the end-to-end lip-reading architecture. The sub-figures in the top row ((a),(b),(c)) refer to the word-level lip-reading training procedures, while the sub-figures in the bottom row ((d),(e),(f)) refer to sentence-level lip-reading. From left to right, (a) and (d) denote training from scratch (the whole model is initialised randomly); (b) and (e) are feature extraction experiments based on visual features extracted from the pre-trained model; and (c) and (f) are fine-tuning experiments. Blue coloured blocks are trained from scratch on the downstream task; yellow coloured blocks are loaded from the pre-trained model and kept frozen during the downstream task; and green coloured blocks are loaded from the pre-trained model and are then fine-tuned for the downstream task. We abbreviate the following model layers: TM: Transformer, FC: Fully-Connected layer, MS-TCN: Multi-Scale Temporal Convolutional Network.} 
\vspace{-10mm}
\label{fig:variations}
\end{table}

\subsection{Datasets}

In this work, we use an unlabelled version of Lip Reading Sentences 3 (LRS3) for pre-training and evaluate the performance of speech representations on LRW and LRS2.
LRW~\cite{DBLP:conf/accv/ChungZ16} is comprised of approximately 500\,000 1.16 second labelled utterances (173 hours in total) featuring a specific word from a 500 word vocabulary. It features hundreds of different speakers recorded in a variety of different backgrounds and head poses.
LRS2~\cite{Afouras18c} is composed of approximately 150\,000 transcribed utterances of varying lengths (224.5 hours in total). This corpus presents a greater challenge since it features a largely unconstrained vocabulary of more than 40\,000 words. Both datasets are collected from BBC programs.

LRS3~\cite{Afouras18d} similarly contains approximately 150\,000 utterances of varying lengths (438.9 hours in total) taken from TED talks. However, these utterances are substantially longer than the ones featured in LRS2, resulting in effectively double the total amount of hours of video and a larger vocabulary. 
This dataset guarantees no overlap between the speakers featured in the train and test sets, meaning that the test set is entirely comprised of speakers that were not seen in other sets.

\vspace{-1mm}
\subsection{Pre-processing}
To crop the mouth Regions of Interest (ROIs), we start by detecting the 68-point facial landmarks using dlib~\cite{dlib09}. We then normalise each frame using a neutral reference frame to remove rotation and size differences. 

Given the transformed facial landmarks, a fixed bounding box is used to crop mouth ROIs with a size of $96\times 96$.

\vspace{-1mm}
\subsection{Data augmentation}
Following~\cite{ma2020towards}, we produce augmented visual streams by applying the techniques of horizontal flipping with a probability of 0.5 and random cropping to a size of $88\times 88$. 
During the testing phase, instead of randomly cropping, we crop a patch of size $88\times 88$ from the centre of the image.
For the word-level classification, mixup with a weight of 0.4 is employed.

\vspace{-1mm}
\subsection{Training settings in the pretext task}
The 3D front-end module preceding our ResNet consists of a convolutional layer with kernel size (5, 7, 7) followed by a max pooling layer. The conformer, on the other hand, is comprised of an initial embedding module -- feed forward layer combined with layer normalisation, dropout (0.1), activation (ReLU -- Rectified Linear Unit) and relative positional encoding (as proposed in~\cite{DBLP:conf/acl/DaiYYCLS19}) -- followed by a set of conformer~\cite{gulati2020conformer} blocks which varies according to the dataset used for the downstream task (6 blocks for LRW, 12 blocks for LRS2).
The conformer blocks feature the following parameters: $d^{{\rm ff}}=2048$, $n^{{\rm head}}=4$, $d^{{\rm q}}=256$, $d^{{\rm k}}=256$, $d^{{\rm v}}=256$; where $d^{{\rm ff}}$ is the hidden dimension of the feed-forward modules, $n^{{\rm head}}$ is the number of self-attention heads, and $d^{{\rm q}}$, $d^{{\rm k}}$, $d^{{\rm v}}$ are the dimensions of the key (K), query (Q), and value (V) in the self-attention layers respectively. The MLP consists of a linear layer with a hidden dimension of 256 units, ReLU activation, dropout, and a linear layer to project the representation to 256-dimensional latent space. For prediction, we average the PASE+ features, which are computed at 100 frames per second (fps), over time to match the frame rate of the input visual features (25 fps). We optimise our model using Adam 
($\beta_{1}=0.9$, $\beta_{2}=0.98$, $\epsilon=10^{-9}$) combined with the Noam scheduler~\cite{DBLP:conf/nips/VaswaniSPUJGKP17} (25\,000 warm-up steps). The model is trained on LRS3 with a batch size of 32. For simplicity, we randomly sample 1 second from each clip and use it as the input to our network, discarding any utterances with less than 1 second in length.

\vspace{-1mm}
\subsection{Training settings in downstream tasks}
\begin{table}[!t]
\renewcommand\thetable{1}
\caption{A comparison of the performance between the baseline methods and ours (pre-trained on LRS3) on the LRW dataset.}
\vspace{-2mm}
\small
\centering
\begin{tabularx}{\columnwidth}{llC}
\toprule
Methods 	&Strategy  & Acc. (\%) \\ \midrule  \midrule
ResNet + BLSTM~\cite{DBLP:conf/interspeech/StafylakisT17} &Supervised & 83.0 \\
Two-stream 3D CNN~\cite{weng2019learning} &Supervised  & 84.1 \\ 
ResNet + BLSTM~\cite{stafylakis2018pushing}&Supervised  & 84.3 \\ 
ResNet + DenseTCN~\cite{ma2020lip}&Supervised   &88.4 \\
PerfectMatch~\cite{9067055} &Self-supervised &71.6 \\
PT-CDDL~\cite{chung2020seeing}&Self-supervised  &75.9 \\
AV-PPC~\cite{DBLP:conf/icpr/TellamekalaVPG20}&Self-supervised  &84.8 \\
\midrule
LiRA-Supervised~\cite{ma2020towards} &Supervised   &87.4 \\
LiRA-Frozen &Self-supervised  &83.1 \\
LiRA-FineTuned &Self-supervised  &88.1\\
\bottomrule
\end{tabularx}
\vspace{-6mm}
\label{tab: lrw_results} 
\end{table}

\noindent\textbf{LiRA-Supervised}\quad
In LiRA-Supervised, we train word-level (Fig.~\hyperref[fig:variations_d]{2a}) and sentence-level lip-reading models (Fig.~\hyperref[fig:variations_d]{2d}) from scratch. In particular, for the task of word-level lip-reading, we add a MS-TCN followed by a linear classifier with an output dimension of 500 on top of the encoder like~\cite{ma2020lip}. A cross-entropy loss is employed to optimise the whole model using Adam with decoupled Weight decay (AdamW)~\cite{loshchilov2019decoupled} with $\beta_{1}=0.9$, $\beta_{2}=0.999$, $\epsilon=10^{-8}$ and a weight decay of 0.01 for 80 epochs with a batch size of 32. The initial learning rate is set to 0.0003. For the task of sentence-level lip-reading, we use 12 multi-head attention blocks ($d^{{\rm ff}}=2048$, $n^{{\rm head}}=4$, $d^{{\rm q}}=256$, $d^{{\rm k}}=256$, $d^{{\rm v}}=256$) together with a linear layer on the top of conformer blocks like~\cite{ma2020large}. Following~\cite{watanabe2017hybrid}, we use a combination of CTC and cross-entropy loss to train a hybrid CTC/Attention architecture for 50 epochs with a batch size of 8. In this case, we use Adam 
with $\beta_{1}=0.9$, $\beta_{2}=0.98$ and $\epsilon=10^{-9}$ with the first 25\,000 steps for warm-up. The initial learning rate is set to 0.0004. At the decoding phase, we use a beam size of 20 for beam search. 
During decoding, we also apply a  transformer-based language model trained on LRS2, LRS3, and Librispeech 960h~\cite{panayotov2015librispeech} (16.2 million words in total). Due to graphic memory limitations, we exclude utterances with more than 600 frames during training.

\noindent\textbf{LiRA-Frozen}\quad
At the end of self-supervised training, the features extracted from the pre-trained frozen encoder are fed to a classifier for evaluation. For word-level lip-reading, we use a MS-TCN, followed by a linear layer with an output size of 500 for classification (Fig.~\hyperref[fig:variations_d]{2b}). For the sentence-level lip-reading, the LiRA features are first fed to 12 conformer blocks, and then the encoded representations are used for CTC/attention joint training (Fig.~\hyperref[fig:variations_d]{2e}).

\noindent\textbf{LiRA-FineTuned}\quad
We follow the same hyperparameter setting as LiRA-Supervised, but instead of training from scratch, we initialise the encoder with the pre-trained weights from the pretext task and then fine-tune the entire model for word-level lip-reading (Fig.~\hyperref[fig:variations_d]{2c}) and sentence-level lip-reading (Fig.~\hyperref[fig:variations_d]{2f}).

\looseness - 1
\section{Results}
\begin{figure}[!t]
    \centering
    \includegraphics[width=\linewidth]{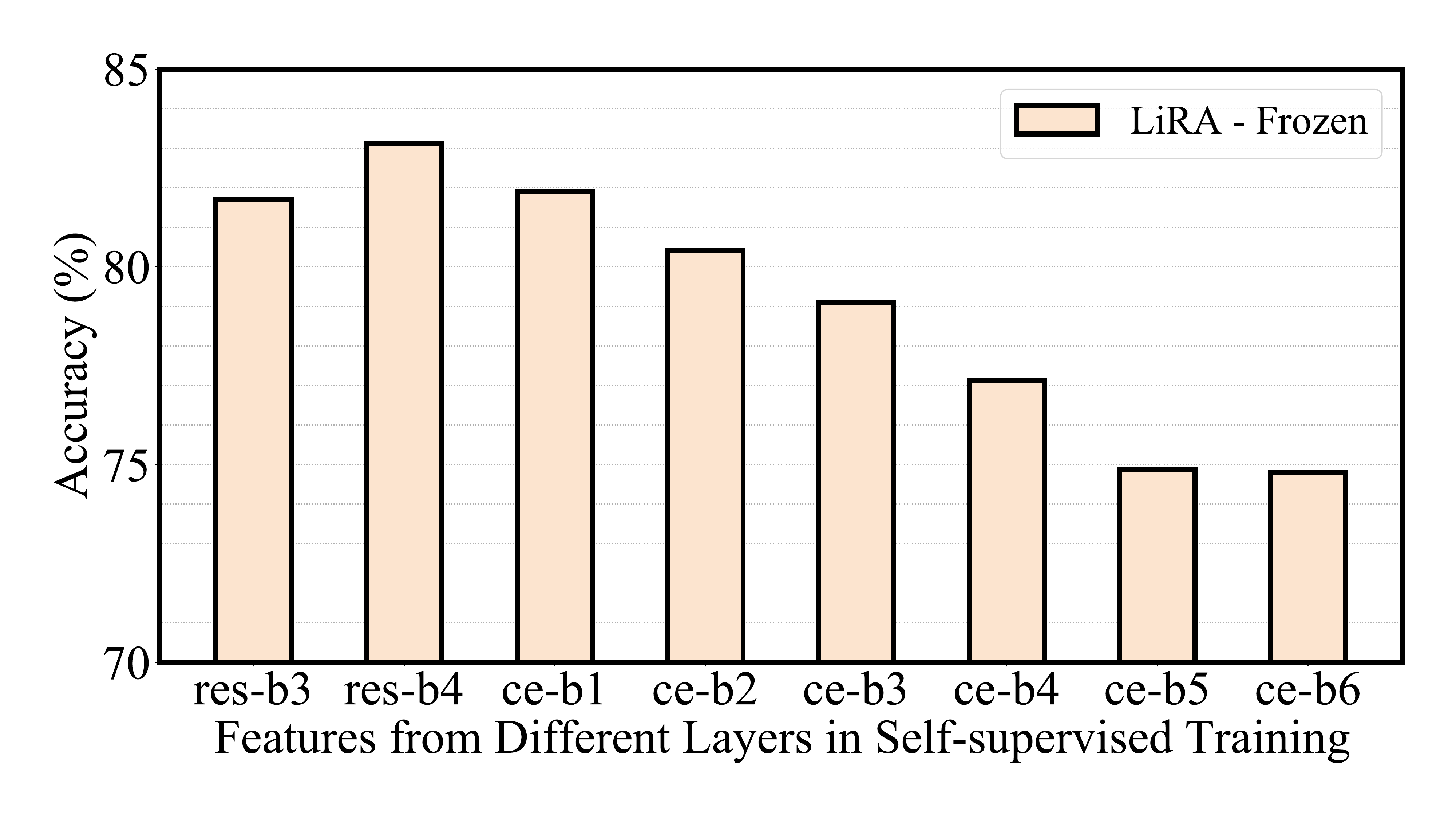}
    \vspace{-8mm}
    \caption{Accuracy of feature classification (LiRA-Frozen) on LRW based on features extracted from different layers after pre-training on LRS3 via self-supervision. ``res-b3'' and ``res-b4'' refer to the output of blocks 3 and 4 from the ResNet-18 respectively; and ``ce-b2'' to ``ce-b12'' refer to the layers from every two conformer blocks from bottom to top.}
    \label{fig:different_positions_on_lrw}
    \vspace{-6mm} 
\end{figure}
\subsection{Word-level lip-reading}
\begin{figure*}[!t]
    \centering
    \begin{subfigure}{.45\textwidth}
    \includegraphics[width=\linewidth]{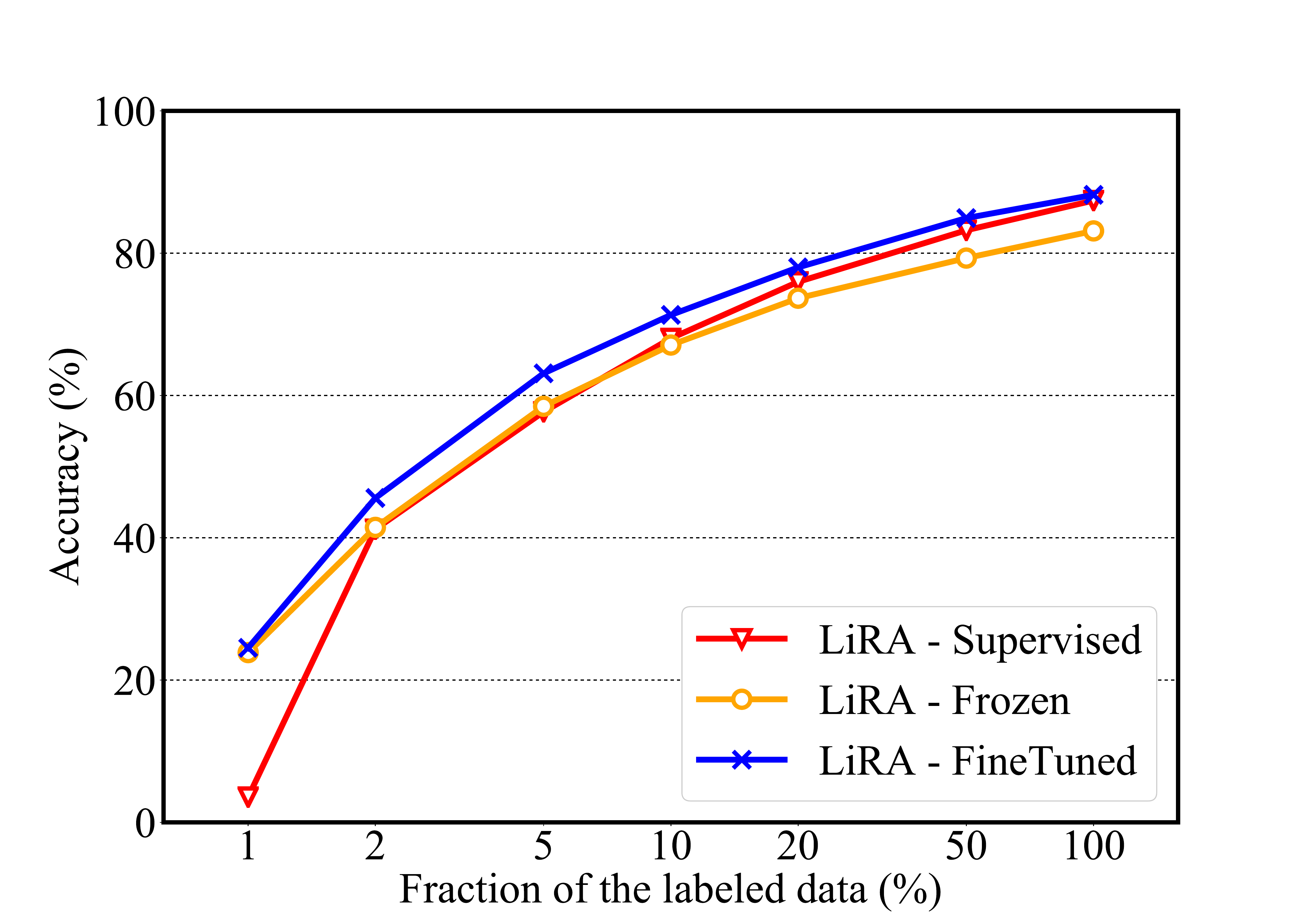}
    \caption{}
    \label{fig:percentage_experiments_on_lrw}  
    \end{subfigure}\hfill
    \begin{subfigure}{.45\textwidth}
    \includegraphics[width=\linewidth]{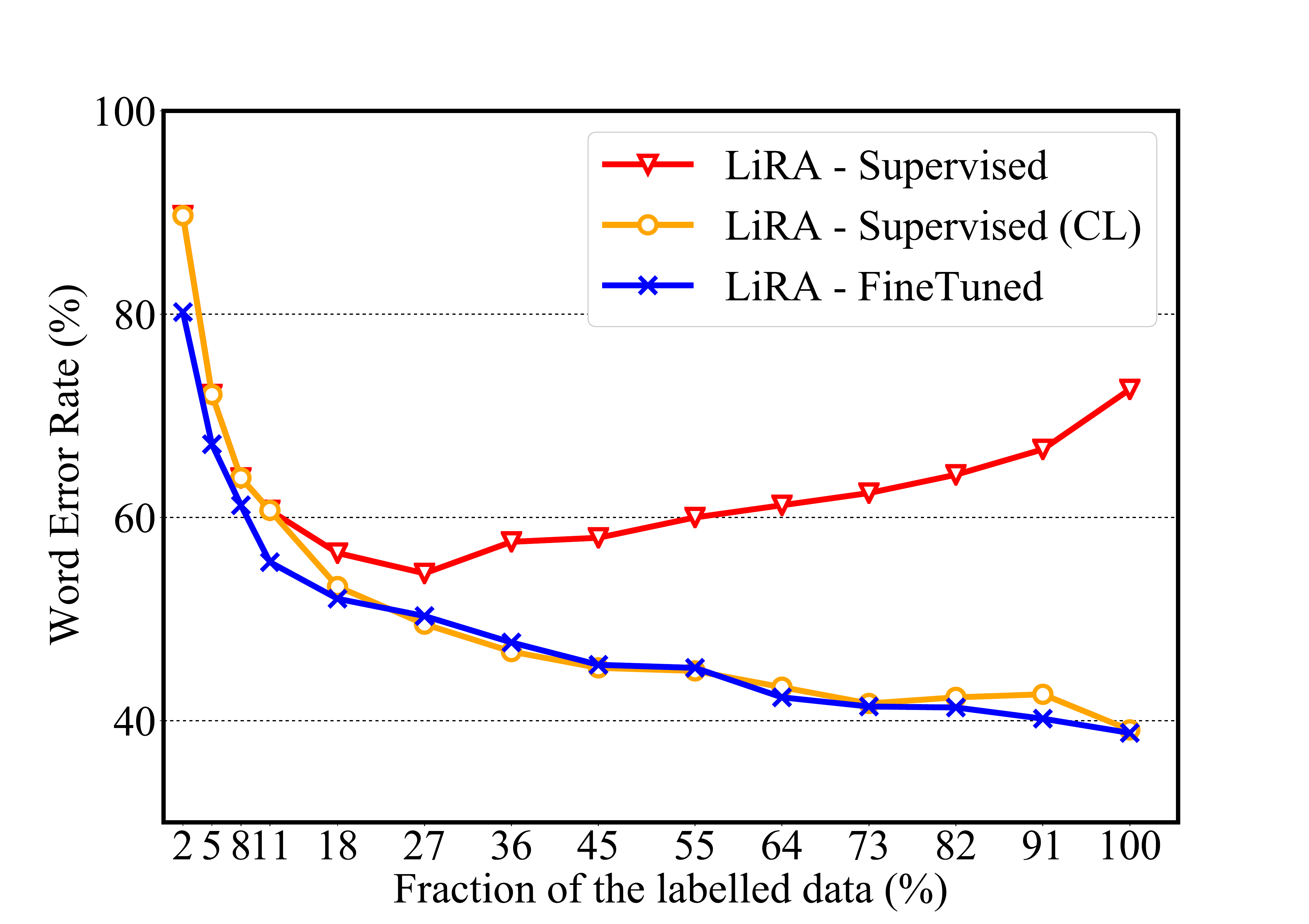}
    \caption{}
    \label{fig:percentage_experiments_on_lrs2}    
    \end{subfigure}
    \vspace{-3mm}
    \caption{Effect of the size of training data on downstream task performance. (a): Accuracy of the end-to-end model as a function of the percentage of the training set (on a logarithmic scale) used for training on LRW. (b) WER achieved by the end-to-end model as a function of the percentage of labelled data used for training on LRS2. All LiRA-Frozen and LiRA-FineTuned models are pre-trained on LRS3 via self-supervision. LiRA-Frozen models are trained using features extracted from the last layer of the ResNet-18 in the pre-trained model, since it achieves the best performance as demonstrated in Fig.~\ref{fig:different_positions_on_lrw}. ``CL'' refers to the model being trained using curriculum learning. LRW and LRS2 contain 165 and 222 hours of labelled training data respectively.}
    \label{fig:percentage_experiments}
    \vspace{-4mm}
\end{figure*}
\vspace{-1mm}

We first evaluate the performance of LiRA-Supervised by training the model from scratch. 
This leads to an accuracy of 87.4\% on LRW which is very close to the state-of-art performance.
For LiRA-Frozen, which is pre-trained on LRS3, the learnt visual speech representations are evaluated on word-level lip-reading by training a MS-TCN classifier on top of the frozen representations, as illustrated in Fig.~\hyperref[fig:variations_b]{2b}. 
Feature extraction performance (LiRA-Frozen) for different layers is portrayed in Fig.~\ref{fig:different_positions_on_lrw}.
We observe that the representations extracted from the last layer of the ResNet-18 achieve a maximum accuracy of 83.1\,\% as seen in Table~\hyperref[tab: lrw_results]{1}. It is clear that the performance generally decreases as the layer becomes deeper, which may indicate that the features extracted in deeper layers are further tuned towards the pretext task and therefore fail to generalise as well for other tasks.

The performance of the 3 downstream scenarios while varying the amount of training data on LRW is shown in Fig.~\ref{fig:percentage_experiments_on_lrw}. We use LRS3 for self-supervised pre-training. We observe that the feature extraction approach leads to superior performance compared to LiRA-Supervised when using smaller fractions of the labelled training set (1-2\,\%). This indicates that the pre-trained model learns useful visual features which also work well on LRW. By adopting this methodology, we can simply train the classification layers while the encoder remains frozen,
and hence significantly reduce the training time of our model. If we fine-tune the full model, including the encoder, then the performance improves further as shown in Fig.~\ref{fig:percentage_experiments_on_lrw}.

\looseness -1
We also observe that the gap between the performance of LiRA-FineTuned and LiRA-Supervised becomes smaller when we increase the amount of labelled data for training. This demonstrates that pre-training using the proposed self-supervised task is particularly beneficial when the labelled training set is very small. In the extreme case, where only 1\,\% of the labelled training data is used, LiRA-Supervised achieves an accuracy of 3.6\,\%. In contrast, we obtain 24.5\,\% accuracy when LiRA-FineTuned is trained using the same amount of data. This is mainly due to the fact that the self-supervised training provides a good initialisation for network training. We also show that LiRA-FineTuned provides an absolute improvement of 0.8\,\% in accuracy over LiRA-Supervised when both are trained on full LRW. This demonstrates that LiRA-FineTuned consistently outperforms LiRA-Supervised, even for larger labelled training sets.

\vspace{-1mm}
\subsection{Sentence-level lip-reading}
\begin{table}[!t]
\renewcommand\thetable{2}
\caption{A comparison of the Word Error Rate (WER) between the baseline methods and ours (pre-trained on LRS3) on the LRS2 dataset. CL: Curriculum learning.}
\vspace{-2mm}
\small
\centering
\begin{tabularx}{\columnwidth}{llC}
\toprule
Methods 	&Strategy  & WER. (\%) \\ \midrule  \midrule
Hyb. CTC/Att.~\cite{petridis2018audio} &Supervised &63.5\\
Conv-seq2seq~\cite{Zhang_2019_ICCV} &Supervised &51.7 \\
TDNN~\cite{yu2020audio} &Supervised &48.9 \\
TM-seq2seq~\cite{Afouras18c} &Supervised &48.3 \\
KD-seq2seq~\cite{DBLP:conf/icassp/AfourasCZ20} &Unsupervised &51.3 \\
\midrule
LiRA-Supervised~\cite{ma2020large} &Supervised (CL) &39.1 \\
LiRA-FineTuned &Self-supervised  &38.8\\
\bottomrule
\end{tabularx}
\vspace{-6mm}
\label{tab: lrs2_results} 
\end{table}

To investigate the performance of visual speech representations in a more challenging task, we run training from scratch (Fig.~\hyperref[fig:variations_d]{2d}) and fine-tuning (Fig.~\hyperref[fig:variations_f]{2f}) experiments on LRS2 after pre-training on LRS3.
We present our results as a function of the fraction of labelled data used during training. 

Results are shown in Fig.~\ref{fig:percentage_experiments_on_lrs2}. It is evident that the performance of LiRA-FineTuned significantly outperforms the supervised baseline. We also observe that the performance of LiRA-Supervised is hard to optimise   without a good initialisation. The performance becomes worse and worse as the training set increases beyond 18\,\% of the total amount of labelled data. This is likely due to the large amount of very long utterances featured in LRS2, which makes training from scratch especially difficult.
To overcome this problem, we use curriculum learning. In particular, we first train the model using 11\,\% of the labelled training set, which is composed of videos with less than 155 frames in length and then use this model for initialisation when training on the entire training set. This curriculum learning strategy results in a substantially more effective training procedure, achieving 39.1\,\% WER for the full dataset. 

Fine-tuning the self-supervised model leads to a small improvement over the curriculum learning strategy resulting
in a 38.8\,\% WER. This is the new state-of-the-art performance on the LRS2 dataset when no external labelled datasets are used for training. We also observe that it leads to a 9.5\,\% absolute improvement compared to the previous state-of-the-art model~\cite{Afouras18c}, as reported in Table~\hyperref[tab: lrs2_results]{2}. Furthermore, as displayed in Fig.~\ref{fig:percentage_experiments}, we are able to outperform the previous state-of-the-art of 48.3\,\% WER using 18$\times$ fewer labelled data -- 76 hours (36\,\% of LRS2) vs 1\,362 hours (MVLRS, LRS2, and LRS3).

\vspace{-1mm}\looseness - 1
\section{Conclusion}
\label{sec:con}
We present LiRA, which learns visual speech representations by cross-modal self-supervised learning. We train a visual model by predicting acoustic features from visual speech, and observe that it can be adapted for lip-reading with remarkable success. By fine-tuning our models for this new task, we achieve an accuracy of 88.1\,\% on LRW and report a WER of 38.8\,\% on LRS2. Given the extent of modern audiovisual corpora, we believe it would be promising to leverage this method towards other visual tasks such as emotion recognition and speaker recognition in the future.

\vspace{-1mm}\looseness - 1
\section{Acknowledgements}
The work of Pingchuan Ma has been partially supported by Honda. The work of Rodrigo Mira has been funded by Samsung. All datasets used in the experiments and all training, testing and ablation studies have been conducted at Imperial College.

\clearpage
\AtNextBibliography{\small}
\section{References}
\begingroup
\setlength\bibitemsep{-.2pt}
\printbibliography[heading=none]
\endgroup

\end{document}